\documentclass[manuscript,screen]{acmart}

\usepackage{listings}
\AtBeginDocument{%
  \providecommand\BibTeX{{%
    \normalfont B\kern-0.5em{\scshape i\kern-0.25em b}\kern-0.8em\TeX}}}

\setcopyright{acmcopyright}
\copyrightyear{2021}
\acmYear{2021}

\acmBooktitle{Biomedical Question Answering: A Survey of Approaches and Challenges}

\begin{document}

\title{Biomedical Question Answering: A Survey of Approaches and Challenges}

\author{Qiao Jin}
\email{jqa14@mails.tsinghua.edu.cn}
\affiliation{
  \institution{Tsinghua University}
  \country{China}
}

\author{Zheng Yuan}
\email{yuanz17@mails.tsinghua.edu.cn}
\affiliation{
  \institution{Tsinghua University}
  \country{China}
}

\author{Guangzhi Xiong}
\email{xgz18@mails.tsinghua.edu.cn}
\affiliation{
  \institution{Tsinghua University}
  \country{China}
}

\author{Qianlan Yu}
\email{yuql18@mails.tsinghua.edu.cn}
\affiliation{
  \institution{Tsinghua University}
  \country{China}
}

\author{Huaiyuan Ying}
\email{yinghy18@mails.tsinghua.edu.cn}
\affiliation{
  \institution{Tsinghua University}
  \country{China}
}

\author{Chuanqi Tan}
\email{chuanqi.tcq@alibaba-inc.com}
\affiliation{
  \institution{Alibaba Group}
  \country{China}
}

\author{Mosha Chen}
\email{chenmosha.cms@alibaba-inc.com}
\affiliation{
  \institution{Alibaba Group}
  \country{China}
}

\author{Songfang Huang}
\email{songfang.hsf@alibaba-inc.com}
\affiliation{
  \institution{Alibaba Group}
  \country{China}
}

\author{Xiaozhong Liu}
\email{liu237@indiana.edu}
\affiliation{
  \institution{Indiana University Bloomington}
  \country{USA}
}

\author{Sheng Yu}
\email{syu@tsinghua.edu.cn}
\affiliation{
  \institution{Tsinghua University}
  \country{China}
}

\renewcommand{\shortauthors}{Jin et al.}
\newcommand{\qiao}[1]{\textcolor{blue}{[#1 -- Qiao]}\typeout{#1}}

\begin{abstract}
Automatic Question Answering (QA) has been successfully applied in various domains such as search engines and chatbots.
Biomedical QA (BQA), as an emerging QA task, enables innovative applications to effectively perceive, access and understand complex biomedical knowledge. 
There have been tremendous developments of BQA in the past two decades,
which we classify into 5 distinctive approaches: classic, information retrieval, machine reading comprehension, knowledge base and question entailment approaches.
In this survey, we introduce available datasets and representative methods of each BQA approach in detail.
Despite the developments, BQA systems are still immature and rarely used in real-life settings.
We identify and characterize several key challenges in BQA that might lead to this issue, and discuss some potential future directions to explore.
\end{abstract}



\maketitle

\section{Introduction} \label{intro}
Biomedical knowledge acquisition is an important task in information retrieval and knowledge management.
Professionals as well as the general public need effective assistance to access, understand and consume complex biomedical concepts.
For example, doctors always want to be aware of up-to-date clinical evidence for the diagnosis and treatment of diseases under the scheme of Evidence-based Medicine \citep{sackett1997evidence}, and the general public is becoming increasingly interested in learning about their own health conditions on the Internet \citep{fox2012online}.

Traditionally, Information Retrieval (IR) systems, such as PubMed, have been used to meet such information needs. 
However, classical IR is still not efficient enough \citep{hersh2002factors, jacquemart2003towards, lee2006beyond, russell2017expert}. 
For instance, \citet{russell2017expert} reported that it requires 4 expert hours to answer complex medical queries using search engines. Compared with the retrieval systems that typically return a list of relevant documents for the users to read, Question Answering (QA) systems that provide direct answers to users' questions are more straightforward and intuitive.
In general, QA itself is a challenging benchmark Natural Language Processing (NLP) task for evaluating the abilities of intelligent systems to understand a question, retrieve and utilize relevant materials and generate its answer.
With the rapid development of computing hardware, modern QA models, especially those based on deep learning \citep{cheng2016long, seo2016bidirectional, chen2017reading, peters2018deep, devlin2019bert}, achieve comparable or even better performance than human on many benchmark datasets \citep{hermann2015teaching, rajpurkar2016squad, joshi2017triviaqa, rajpurkar2018know, yang2018hotpotqa} and have been successfully adopted in general domain search engines and conversational assistants \citep{qiu2017alime, zhou2020design}.

The Text REtrieval Conference (TREC) QA Track has triggered the modern QA research \citep{voorhees2001trec}, when QA models were mostly based on IR.
\citet{zweigenbaum2003question} first identified the distinctive characteristics of BQA over general domain QA.
Later, many classic BQA systems have been proposed, such as EPoCare \citep{niu2003answering}, PICO-(P: patient/problem, I: intervention, C: comparison, O: outcome) and knowledge-extraction-based BQA systems \citep{demner2005knowledge, demner2006answer, demner2007answering}, MedQA \citep{yu2007development}, \citet{terol2007knowledge}, \citet{weiming2007automatic}, Health on the Net QA (HONQA) \citep{cruchet2008supervised}, AskHERMES \citep{cao2011askhermes} etc.
Such systems employ complex pipelines with numerous question, document and answer processing modules, which is typically reflected by the IBM Watson system \citep{ferrucci2012watson}.
BioASQ \citep{tsatsaronis2015overview} is a cornerstone challenge that has been running annually since 2013 for the evaluation of biomedical natural language understanding systems.
A variety of BQA systems have been proposed in BioASQ, improving QA performance from approximately 20\% top factoid mean reciprocal rank and list F-measure in BioASQ 1 to approximately 50\% in BioASQ 8 \citep{Nentidis2020overview}.
Notably, the landscape of BioASQ participating models has been re-shaped by several NLP methodological revolutions:
1. The introduction of distributed word representations \citep{mikolov2013efficient, mikolov2013distributed};
2. Deep learning-based QA models such as Bi-Directional Attention Flow (Bi-DAF) \citep{seo2016bidirectional};
3. Large-scale Pre-trained Language Models (PLMs) represented by Embeddings for Language Models (ELMo) \citep{peters2018deep} and bidirectional encoder representations from transformers (BERT) \citep{devlin2019bert}.
Currently, almost all top-performing BQA systems use the biomedical PLMs (e.g. BioBERT \citep{lee2020biobert}), in their systems.
Furthermore, various other BQA challenges and datasets have been introduced to further facilitate BQA research in different directions, e.g.: LiveQA-Med \citep{abacha2017overview}, MEDIQA \citep{abacha2019overview, savery2020question} for consumer health, emrQA \citep{pampari2018emrqa} for clinical BQA, VQA-Rad \citep{lau2018dataset}, VQA-Med \citep{abacha2019vqa} and PathVQA \citep{he2020pathvqa} for visual BQA.

Despite the tremendous developments, BQA is still immature and faces several key challenges:
\begin{itemize}
\item \textbf{Dataset Scale, Annotation \& Difficulty}: 
Most current BQA models utilize deep learning and are thus data-hungry. 
However, annotating large-scale biomedical corpora or knowledge bases is prohibitively expensive. 
As a result, current expert-annotated BQA datasets are small in size, with only hundreds to few thousand QA instances.
To build large-scale datasets, many works have attempted to automatically collect BQA datasets, but their utility is limited and their annotation quality is not guaranteed.
Furthermore, questions of most current BQA datasets do not require complex reasoning to answer.
\item \textbf{Domain Knowledge Not Fully Utilized}:
There are rich biomedical resources that encapsulate different types of domain knowledge, including large-scale corpora, various biomedical KBs, domain-specific NLP tools and images.
Unfortunately, most BQA models fail to utilize them effectively.
As a result, biomedical domain knowledge is not fully utilized, which can be potentially solved by fusing different BQA approaches. 
\item \textbf{Lack of Explainability}:
Since biomedicine is a highly specialized domain, ideal systems should not only return the exact answers (e.g.: ``yes"/``no"), but also provide the explanations for giving such answers.
However, there are only a few BQA systems that are explainable.
\item \textbf{Evaluation Issues}:
Qualitatively, current evaluations mainly focus on certain modules, e.g. Machine Reading Comprehension (MRC), within a complete QA pipeline.
Quantitatively, most evaluation metrics do not consider rich biomedically synonymous relationships.
\item \textbf{Fairness and Bias}:
Most machine-learning-based BQA systems learn from historical data, such as scientific literature and electronic medical records, which 
can be potentially biased and out of date.
However, studies on BQA fairness and model transparency are quite sparse. 
\end{itemize}

This paper is organized as follows:
We first describe the scope of this survey in \S\ref{scope};
We then give an overview of the surveyed BQA approaches in \S\ref{approach}. 
Various methods and datasets have been proposed for each BQA approach, and they are systematically discussed in \S4-8;
To conclude, we summarize several challenges of BQA and discuss potential future directions in \S\ref{challenge}.


\section{Survey Scope} \label{scope}
Biomedicine is a broad domain that covers a range of biological and medical sciences.
Since the term is often loosely used by the community, we specifically define several sub-domains of biomedicine, namely scientific, clinical, consumer health and examination, as the focus of this survey.
Each content type is defined by the most distinctive characteristics of their corresponding users, questions and expected answers, as shown in Table \ref{tab:content_type}.
It should be noted that the content types are not mutually exclusive, but most of our surveyed works belong to only one of them.
In this section, we introduce these contents in \S2.1-2.4, and we also describe some related surveys with the focus of different scopes in \S\ref{related_surveys}.
Several typical QA examples for each content type are shown in Table \ref{tab:content_eg}.
The datasets are selected from hand literature search on PubMed and Google Scholar with keywords such as “biomedical”, “biological”, “medical”, “clinical”, “health” and “question answering”. For each included dataset paper, we also checked their references and papers citing it. 
We describe mostly the methods with state-of-the-art performance on the surveyed datasets.

\begin{table*}[ht!]
    \small
    \centering
    \begin{tabular}{p{2.8cm}p{2.5cm}p{5cm}p{3.2cm}}
    \toprule
    \textbf{Content} & \textbf{Main User} & \textbf{Question motivation} & \textbf{Answer style} \\
    \midrule
    Scientific (\S\ref{scientific}) & -- & Learning cutting-edge scientific advances & Professional-level \\
    \midrule
    Clinical (\S\ref{clinical}) & Professionals & Assisting clinical decision making & Professional-level \\
    \midrule
    Consumer health (\S\ref{consumer}) & General public & Seeking advice or knowledge & Consumer-understandable \\
    \midrule
    Examination (\S\ref{examination}) & -- & Testing biomedical knowledge & Mostly choices \\
    \bottomrule
    \end{tabular}
    \caption{Characteristics of different BQA contents. `--' denotes no specific users.}
    \label{tab:content_type}
\end{table*}

\begin{table*}
    \small
    \centering
    \begin{tabular}{p{2cm}p{4cm}p{4cm}p{3.5cm}}
    \toprule
    \textbf{Type / Dataset} & \textbf{Question} & \textbf{Context} & \textbf{Answer} \\
    \midrule
    \textbf{Scientific} \\
    \midrule
    BioASQ & Is the protein Papilin secreted? & [...] secreted extracellular matrix proteins, mig-6/papilin [...] & Yes \\
    \midrule
    Biomed-Cloze & Helicases are motor proteins that unwind double stranded \underline{   ?   } into [...] & Defects in helicase function have been associated with [...] & nucleic acid  \\ 
    \midrule
    \textbf{Clinical} \\
    \midrule
    emrQA & Has the patient ever had an abnormal BMI? & 08/31/96 [...] BMI: 33.4 Obese, high risk. Pulse: 60. resp. rate: 18 & BMI: 33.4 Obese, high risk \\
    \midrule
    CliCR & If steroids are used , great caution should be exercised on their gradual tapering to avoid \underline{   ?   } & [...] Thereafter, tapering of corticosteroids was initiated with no clinical relapse. [...] & relapse \\
    \midrule
    \textbf{Consumer} \\
    \midrule
    MedQuAD & Who is at risk for Langerhans Cell Histiocytosis? & NA & Anything that increases your risk of [...] \\
    \midrule
    MEDIQA-AnS & What is the consensus of medical doctors as to whether asthma can be cured? And do you have [...] & Asthma Overview Asthma is a chronic lung disease that causes episodes of wheezing [...] & Asthma is a chronic disease. This means that it can be treated but not cured. [...] \\
    \midrule
    \textbf{Examination} \\
    \midrule
    HEAD-QA & The antibiotic treatment of choice for [...] is & 1. Gentamicin; 2. Erythromycin; 3. Ciprofloxacin; 4. Cefotaxime & 4. Cefotaxime \\
    \bottomrule
    \end{tabular}
    \caption{Typical question-answer examples of different content types.}
    \label{tab:content_eg}
\end{table*}

\subsection{Scientific} \label{scientific}
Scientific QA addresses cutting-edge questions whose answers need to be extracted or inferred from scientific literature, e.g.: ``Which cells express G protein-coupled receptors?".
Most of the new findings in the biomedical field are published in the form of scientific literature, whose size is growing at an unprecedented pace:
for example, MEDLINE\footnote{\url{https://www.nlm.nih.gov/bsd/medline.html}}, a bibliographic database of life sciences, contains references to over $30$M articles and about $2.7$k articles are added each day in 2019.
Given the huge number of scientific publications, 
it's almost impossible to manually read all relevant studies and give comprehensive answers to scientific questions, so automatic answering of scientific questions is vital.

The BQA community's fight against COVID-19 is a great example of scientific QA.
There has been a surge of COVID-19-related publications \citep{li2020surging} that human experts find difficult to keep up with.
Consequently, it's important to develop automatic methods for natural language understanding of them. 
To facilitate NLP studies on the COVID-19 literature, \citet{wang2020cord} release the CORD-19 corpus which contains more than 280k papers about the novel coronavirus.
Many BQA datasets have been introduced to help develop and evaluate models that answer COVID-19-related questions, e.g.: COVID-QA and COVIDQA datasets and the EPIC-QA challenge.
Several resources and methods \citep{sun2020analysis, poliak2020collecting, zhang2020cough} have been introduced to tackle the COVID-19 QA by the QE approach (\S\ref{qeapproach}).

The most distinctive feature of scientific BQA is that large-scale corpora like PubMed and PubMed Central are freely available, which contain $4.5$B and $13.5$B tokens, respectively.
In contrast, the entire English Wikipedia contains only $2.5$B tokens.
Besides, documents in PubMed and PubMed Central are semi-structured -- they have sections of background, introduction, methods, conclusion etc., which can be potentially exploited in building domain-specific datasets.
Consequently, the largest expert-annotated BQA dataset -- BioASQ, and most large-scale (semi-)automtically constructed BQA datasets are all scientific BQA datasets (discussed in \S\ref{automatic}).
Further exploiting the scale and structure of the scientific literature to design novel BQA tasks remains an interesting direction to explore.

\subsection{Clinical} \label{clinical}
Clinical QA focuses on answering healthcare professionals' questions about medical decision making for patients.
\citet{ely2000taxonomy} find the most frequent clinical questions are:
1. What is the drug of choice for condition x? ($11$\%);
2. What is the cause of symptom x? ($8$\%);
3. What test is indicated in situation x? ($8$\%);
4. What is the dose of drug x? ($7$\%);
5. How should I treat condition x (not limited to drug treatment)? ($6$\%);
6. How should I manage condition x (not specifying diagnostic or therapeutic)? ($5$\%);
7. What is the cause of physical finding x? ($5$\%);
8. What is the cause of test finding x? ($5$\%);
9. Can drug x cause (adverse) finding y? ($4$\%);
10. Could this patient have condition x? ($4$\%).

Most of the clinical questions shown above are generic (case 1-9) and largely non-specific to patients.
In this case, clinical QA is similar to consumer health QA (\S\ref{consumer}).
If the questions are specific to certain patients (e.g.: case 10), their Electronic Medical Records (EMRs) should be provided. 
EMRs store all health-related data of each patient in both structured (i.e.: tables) and unstructured (i.e.: medical notes) formats.
Due to the complexity and size of the EMR data, it's time-consuming and ineffective for the doctors to manually check the EMRs for clinical questions about the patient.
Clinical QA systems can meet such information needs by quickly and accurately answering these questions.
The difficulty of clinical BQA largely lies in the annotation of QA pairs, where considerable medical expertise and reasoning across clinical notes should be required to answer the questions \citep{raghavan2018annotating}.
For this, \citet{pampari2018emrqa} use expert-annotated templates (e.g.: ``Has the patient ever been on \{medication\}?") with the existing i2b2 dataset annotations\footnote{\url{https://www.i2b2.org/NLP/DataSets/}} (e.g.: ``[...] Flagyl $<$medication$>$ [...]") to build the first large-scale EMR BQA dataset emrQA.
\citet{yue2020clinical} analyze the emrQA dataset and find: 1. the answers are usually incomplete; 2. the questions are often answerable without using domain knowledge.
Both are caused by the dataset collection approach of emrQA.
Another large-scale clinical QA dataset, CliCR \citep{suster2018clicr}, is built by cloze generation (\S\ref{automatic}).

\citet{roberts2017semantic} show that the structured information of EMRs can be effectively queried by semantic parsing, where the goal is to map the natural language questions to their logic forms \citep{kamath2018survey}, e.g.: Q: ``Was her ankle sprain healed?" Logic form: is\_healed(latest(lambda(ankle sprain))).
To tackle the clinical QA of structured EMR data, \citet{soni2019using} annotate a dataset of $1$k clinical questions with their logic forms.
Some paraphrasing-based data augmentation methods are also introduced to improve the performance of semantic parsers of EMR questions \citep{soni2019paraphrase, soni2020paraphrasing}.
\citet{wang2020text} propose TREQS, a two-stage generation model based on the sequence-to-sequence model and the attentive-copying mechanism, and show its effectiveness on their MIMICSQL dataset for the question-to-SQL (table-based) task.
Based on MIMICSQL dataset, \citet{park2020knowledge} propose a question-to-SPARQL (graph-based) dataset: MIMIC-SPARQL$*$. 
TREQS also performs better on the graph-based dataset.

Radiology and pathology images play a vital role in the diagnosis and treatment of diseases.
Clinical QA also contains VQA tasks, e.g.: VQA-Rad \citep{lau2018dataset}, VQA-Med \citep{abacha2019vqa} and PathVQA \citep{he2020pathvqa}, which help doctors to analyze a large amount of images required for medical decision making and population screening.

\citet{ely2005answering} also study the obstacles that prevent physicians from answering their clinical questions, and find that doubting whether the answer exists is the most commonly (11\%) reported reason for not pursuing the answers and the most common obstacle in pursuing the answer is the failure to find the needed information in the selected resources (26\%).
Both problems can be solved by the clinical QA system.
Currently, the main challenge for building such systems is the lack of large-scale expert-annotated datasets that reflect the real demands in the clinic.
Apart from the high-price of deriving such annotations, there are also privacy and ethical issues for releasing them, especially when the datasets are based on EMRs.
Future clinical QA datasets should have larger scales, less noise and more diversity.

\subsection{Consumer Health} \label{consumer}
Consumer health questions are typically raised by the general public on search engines,
where online medical services provide people with great convenience as they are not limited by time and space.
As a result, rapidly increasing numbers of consumers are asking health-related questions on the Internet:
According to one report released by the Pew Research Center \citep{fox2012online}, over one-third of American adults have searched online for medical conditions that they might have.
Many try to find answers to their medical questions before going to a doctor or making decisions about whether to go to a doctor,
and their information needs range from self-diagnosis to finding medications.
It is vitally important to provide accurate answers for such questions,
because consumers are unable to judge the quality of medical contents.
Considering the contradiction between the great demands of consumers and the scarcity of medical experts, an automatic answering system is helpful for sharing medical resources to provide online medical service.

Some works \citep{zhang2017chinese, zhang2018multi,tian2019chimed} have exploited the doctors' answers to patients' questions on online medical consultation websites e.g.: XunYiWenYao\footnote{\url{http://xywy.com/}}, to build large-scale consumer health QA datasets.
These datasets are formatted as multi-choice BQA, where the task is to find the relevant or adopted answers.
However, the quality of such datasets is questionable since the answers are written by users from online communities and the forum data has intrinsic noise.
Remarkable efforts have been made by NLM's Consumer Health Information and Question Answering (CHIQA) project\footnote{\url{https://lhncbc.nlm.nih.gov/project/consumer-health-question-answering}}.
CHIQA \citep{demner2020consumer} is aimed at dealing with a vast number of consumer requests (over 90k per year) by automatically classifying the requests and answering their questions.
It also provides various datasets to develop consumer health BQA methods, including question decomposition and type, named entity and spelling error datasets.

For consumer health QA, understanding the questions of consumers is a vital but difficult step: such questions might contain many spelling and grammar errors, non-standard medical terms and multiple focuses \citep{roberts2016interactive,zhang2017chinese}. 
For example, \citet{abacha2019summarization} find that consumers often submit long and complex questions that lead to substantial false positives in answer retrieval.
To tackle it, they introduce the MeQSum corpus\footnote{\url{https://github.com/abachaa/MeQSum}} that contains $1$k summarized consumer health questions and achieve the best $44.16$\% ROUGE-1 score using pointer-generator network \citep{see2017get} with semantic augmentation from question datasets.
On the other hand, most consumers have no biomedical domain knowledge, so the returned answers should be not only accurate but also explainable (\S\ref{explainability}), posing further challenges for consumer health QA.

\subsection{Examination} \label{examination}
Many general domain QA datasets that are extracted from examinations have been introduced \citep{shibuki2014overview, penas2014overview, khashabi2016question, lai2017race}.
Similarly, Examination BQA has that addresses automatic answering of medical examination questions also been explored.
For example, in many countries, medical licensure requires the passing of specific examinations, e.g.: USMLE\footnote{\url{https://www.usmle.org/}} in the US.
Test items in examinations often take the form of multi-choice questions,
and answering them requires comprehensive biomedical knowledge.
Several datasets have been released that exploit such naturally existing QA data, e.g.: HEAD-QA \citep{vilares2019head} and NLPEC \citep{li2020towards}.
Usually, no contexts are provided for such questions and automatic answering of them requires the systems to find supporting materials (e.g.: texts, images and KBs) as well as reason over them.
However, the real utility of examination QA is still questionable.

\subsection{Related Surveys} \label{related_surveys}
\citet{athenikos2010biomedical} systematically review BQA systems, mainly classic ones published before 2010.
Content-wise, they classify BQA into biological QA and medical QA, which roughly correspond to our scientific and clinical content types, respectively.
\citet{bauer2012usability} and \citet{sharma2015survey} briefly compare several classic BQA systems.
\citet{neves2015question} present a detailed survey of QA for biology, which we classify as scientific BQA in this paper.
This survey also discusses various biological BQA systems.
Recently, \citet{nguyen2019question} identifies several challenges in consumer health BQA, including the lack of publicly available datasets, term ambiguity, incontinuous answer spans and the lack of BQA systems for patients.
\citet{nguyen2019question} proposes a research plan to build a BQA system for consumer self-diagnosis.
\citet{kaddari2020biomedical} presents a brief survey that discusses several scientific BQA datasets and methods.

\section{BQA Approach Overview} \label{approach}
\begin{figure*}[h]
    \centering
    \includegraphics[width=\linewidth]{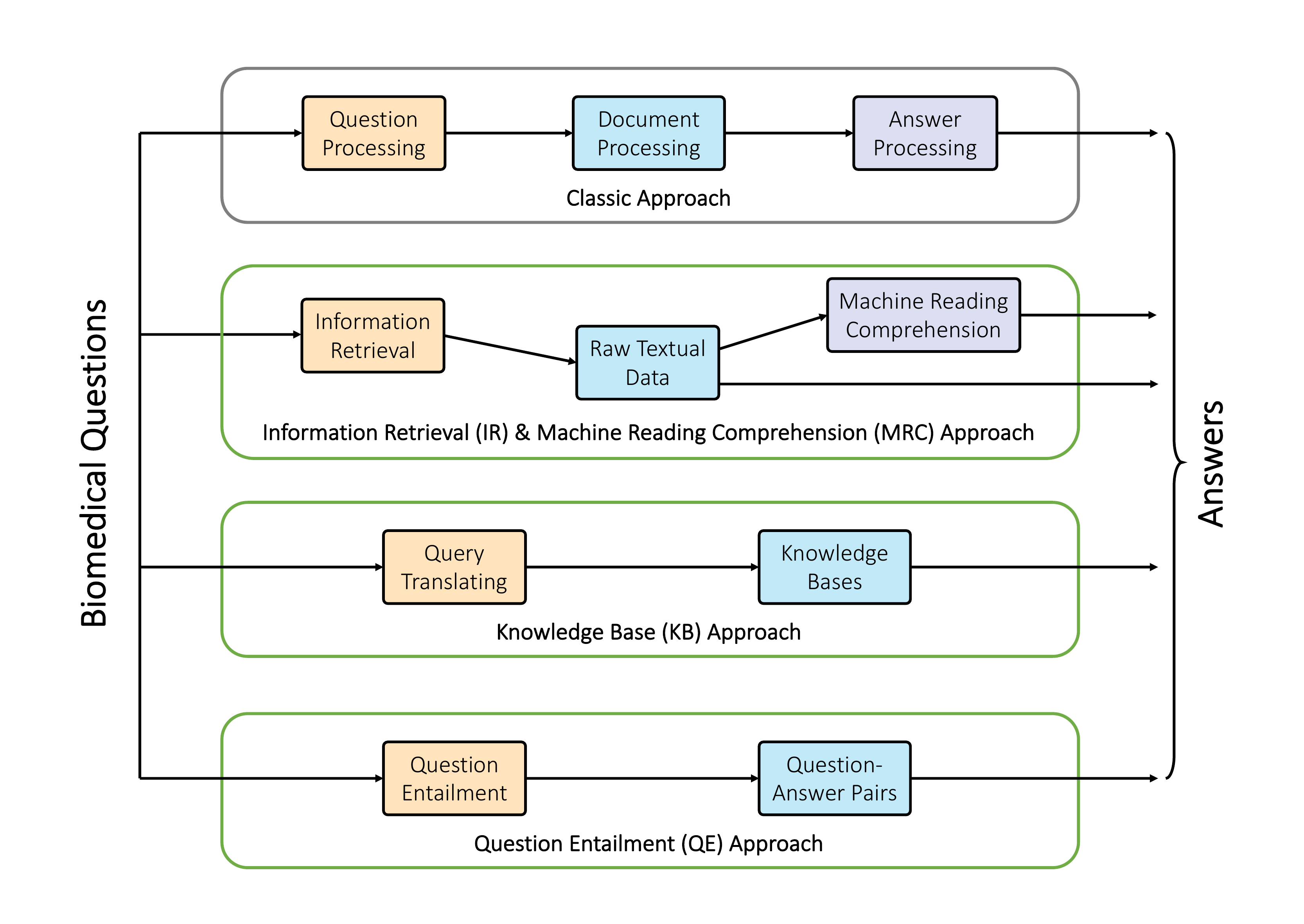}
    \caption{Overview of major biomedical question answering approaches. Boxes indicate methods or resources.}
    \label{fig:arch}
\end{figure*}
\noindent The fundamental task of BQA is to answer a given questions about biomedicine.
In this survey, each BQA approach denotes a distinctive means to tackle the task.
We briefly define different BQA approaches in this section, and a high-level overview of them is shown in Figure \ref{fig:arch}.

We first define the \textbf{Classic BQA approach} from a historical perspective.
Mainly due to the lack of modular QA datasets (e.g. MRC BQA datasets), systems of this approach typically: 
1. contain many sub-tasks and follow the pipeline of the question, document and answer processing, similar to IBM's Watson system \citep{ferrucci2012watson}; 
2. use many rule-based and/or sparse-feature-based machine learning modules in one system; 
3. are evaluated on small-scale private datasets.
Since most of the classic BQA systems are surveyed in detail by \citet{athenikos2010biomedical}, we just briefly introduce them in \S\ref{classic}.

Besides the classic BQA approach, other BQA approaches tackle the task using specific supporting resources that are included in standard, public datasets.
They typically contains a collection of datasets and methods, and we define several BQA approaches below:
\begin{itemize}
\item \textbf{Information Retrieval (IR) approach}: where systems retrieve relevant documents to answer the questions;
\item \textbf{Machine Reading Comprehension (MRC) approach}: where systems read given contexts about the questions to predict the answers. The contexts of MRC approach can be provided by the IR approach;
\item \textbf{Knowledge Base (KB) approach}: where systems either explicitly translate the input questions to RDF queries to search the KBs or implicitly use the integrated knowledge from certain biomedical KBs to get the answers;
\item \textbf{Question Entailment (QE) approach}: where systems find similar questions that have been previously answered in a Q-A pair database and re-use their answers to answer the given question;
\end{itemize}
Characteristics of these approaches are summarized in Table \ref{tab:approach_type}.

\begin{table*}[ht!]
    \small
    \centering
    \begin{tabular}{p{4cm}p{4.5cm}p{4.2cm}}
    \toprule
    \textbf{BQA Approach} & \textbf{Supporting resources} & \textbf{Answer form} \\
    \midrule
    IR BQA approach (\S\ref{irbqa}) & Document collections & Specific documents \\
    \midrule
    MRC BQA approach (\S\ref{mrcbqa}) & Specific documents (contexts) & Y/N; Extraction; Generation \\
    \midrule
    KB BQA approach (\S\ref{kb}) & Knowledge bases & Biomedical entities/relations \\
    \midrule
    QE BQA approach (\S\ref{qeapproach}) & Answered questions (FAQs) & Existing answers of similar questions \\
    \bottomrule
    \end{tabular}
    \caption{Characteristics of different BQA approaches. Y/N: Yes/No.}
    \label{tab:approach_type}
\end{table*}


\section{Classic BQA} \label{classic}
In this section, We breifly introduce several representative classic BQA systems, and point the readers to the BQA survey by \citet{athenikos2010biomedical} for more details.

\begin{table*}[ht!]
    \small
    \centering
    \begin{tabular}{p{2.2cm}p{1cm}p{3cm}p{3.6cm}p{3.2cm}}
    \toprule
    \textbf{System} & \textbf{Content} &  \textbf{Question Processing} & \textbf{Document Processing} & \textbf{Answer Processing} \\ \midrule
    EPoCare \citep{niu2003answering} & Clinical & PICO format & Keyword-based retriever & -- \\ \midrule
    \citet{takahashi2004question} & Scientific & Question type classification; Query formulation & MySQL retriever & -- \\ \midrule
    \citet{demner2005knowledge, demner2006answer, demner2007answering} & Clinical & PICO-format;  Query formulation& Knowledge extraction;  Semantic matching& Semantic clustering  \& summarization \\ \midrule
    MedQA  \citep{lee2006beyond, yu2007development}& Consumer & Question type  classification & Lucene retriever &  Answer extraction  \& summarization \\ \midrule
    BioSquash  \citep{shi2007question}& Scientific & Semantic annotation & Semantic annotation; Graph construction& Sentence selection,  clustering and  post-processing \\ \midrule
    \citet{terol2007knowledge} & Clinical & Question and  answer type  classification;  Logic form  extraction&  -- & Answer generation  based on logic forms \\ \midrule
    \citet{weiming2007automatic} & Clinical & Concept and  relation recognition& Lucene retriever & Semantic interpretation  and clustering  based on relations \\ \midrule
    HONQA  \citep{cruchet2008supervised}& Consumer  & Question, expected  answer and medical  type classification& -- & --  \\ \midrule
    \citet{lin2008biological} & Scientific & Question type  classification;  Query expansion& Google-interfacing  retriever& NER- \& SRL-based \\ \midrule
    EAGLi  \citep{gobeill2009question}& Scientific  & Query and target  categorization& PubMed e-utils  and EasyIR  \citep{aronson2005fusion}& Extracting  concepts  \\ \midrule
    AskHERMES  \citep{cao2011askhermes}& Clinical & Topic classification  with MetaMap  \citep{yu2008automatically}& BM25 retriever; Longest common subsequence extractor & Content Clustering \\ \midrule
    MiPACQ  \citep{cairns2011mipacq}& Clinical  & Semantic annotation& Lucene retriever & ML-based re-ranking  \\ 
    \bottomrule
    \end{tabular}
    \caption{An overview of classic BQA systems (listed in chronological order). \text{``--"}: no special processing steps.}
    \label{tab:traditional}
\end{table*}

Traditionally, QA approaches consist of 3 main parts \citep{hirschman2001natural}: 
1. Question processing, where systems (a) determine the type of the question and the corresponding type of the expected answer and (b) form queries that are fed to certain document retrieval systems;
2. Document processing, where systems (a) retrieve relevant documents from the queries generated in the previous step and (b) extract answer candidates from the relevant documents;
3. Answer processing, where systems rank the candidate answers based on certain criteria.
Although recently some of these modules have become independent QA approaches (e.g. IR, MRC), classic BQA still remains a distinctive class of approach in this survey because most of these works describe a whole system that includes all these subtasks.
We show an overview of classic BQA methods in Table \ref{tab:traditional} with their specific methods for question, document and answer processing. 

\paragraph{\textbf{PICO-based:}}
\citet{niu2003answering} explore PICO-based BQA in the EPoCare project using simple keyword-based retrievers.
Demner-Fushman and Lin further study the PICO-based semantic BQA for the practice of Evidence-Based Medicine (EBM) in a series of works \citep{demner2005knowledge, demner2006answer, demner2007answering}, where the core step involves searching PubMed articles that are annotated with extracted medical knowledge.
\citet{huang2006evaluation} study the feasibility of using the PICO format to represent clinical questions and conclude that PICO is primarily focused on therapy type clinical questions and unsuitable for representing the others (e.g.: prognosis, etiology).

\paragraph{\textbf{Natural-language-based:}}
To tackle a broader range of topics, most other classic BQA systems accept natural language questions:
The medical definitional question answering system (MedQA, \citet{lee2006beyond, yu2007development}) is the first fully implemented BQA system that generates answers by extractive summarization for users' definitional questions from large text corpora.
BioSquash \citep{shi2007question} is adapted from the general domain summarizer Squash \citep{melli2005description} and is focused on QA-oriented summarization of biomedical documents.
\citet{terol2007knowledge} utilize logic forms for BQA, where the core step is to derive the logic forms of questions and utilize them to generate answers.
HONQA \citep{cruchet2008supervised} is a French/English bilingual BQA system that focuses on the supervised classification of question and answer types for question answering.
\citet{lin2008biological} explore answering biomolecular event questions with named entities using syntactic and semantic feature matching.
\citet{gobeill2009question} generate 200 questions from biomedical relational databases to evaluate their EAGLi platform.
\citet{cao2011askhermes} propose the askHERMES system, a BQA system that performs several semantic analyses, including question topic classification and content clustering, to provide extractive summaries for clinical questions.

The classic BQA approaches rely heavily on rule-based models and various ad-hoc modules in their complex pipelines.
Although these might be necessary in industry-level applications, they are hard to develop and maintain in academic settings.
In addition, most classic BQA systems have not been validated on large-scale public datasets.
With the introduction of various BQA datasets that are focused on specific BQA topics or steps, only a few BQA systems that tackle the complete question-to-answer BQA task have been proposed recently, which will be discussed as the modular evaluation issue in \S\ref{eval}.

\section{Information Retrieval BQA} \label{irbqa}
Information Retrieval (IR) BQA denotes the approach that uses \textit{IR BQA Methods} to retrieve relevant text snippets from certain \textit{Document Collections} for the given question, where the retrieved snippets can be either directly used as answers or further fed to MRC models (\S \ref{mrcbqa}).
We also discuss several \textit{IR BQA Datasets} that are summarized in Table \ref{tab:irbqa}.

\begin{table*}[ht!]
    \small
    \centering
    \begin{tabular}{p{3.5cm}p{1.8cm}p{1cm}p{4cm}p{1.5cm}p{1.2cm}}
    \toprule
    \textbf{Dataset} & \textbf{Size} & \textbf{Metric} & \textbf{State-of-the-art (\%)} & \textbf{Content} & \textbf{Format}
    \\  \midrule
    BioASQ Task B Phase A  \citep{tsatsaronis2015overview} & $3.7$k & MAP & 33.04 (document) / 68.21 (snippet)$^1$ & Scientific & Retrieval  \\ \midrule
    BiQA  \citep{lamurias2020generating} & $7.4$k & MAP & -- & Consumer & Retrieval  \\ \midrule
    EPIC-QA$^2$ & 45 & MAP & -- & Sci. \& Con. & Retrieval  \\ \midrule
    HealthQA  \citep{zhu2019hierarchical} & $7.5$k & MRR & 87.88  \citep{zhu2019hierarchical} & Consumer & Retrieval  \\ \midrule
    TREC Genomics  \citep{hersh2006trec, hersh2007trec} & $28$ (06), $36$ (07) & MAP & 54.39 (06) / 32.86 (07)  \citep{hersh2009trec} & Scientific & Retrieval  \\
    \bottomrule
    \end{tabular}
    \caption{An overview of the information retrieval biomedical question answering datasets (listed in alphabetical order). $^1$Batch 2 of BioASQ Task 8 b Phase A. $^2$\url{https://bionlp.nlm.nih.gov/epic_qa/}}
    \label{tab:irbqa}
\end{table*}

\subsection{Document Collections}
PubMed\footnote{\url{https://pubmed.ncbi.nlm.nih.gov/}} and PubMed Central\footnote{\url{https://www.ncbi.nlm.nih.gov/pmc/}} are the most widely used corpora.
Both were developed and are maintained by the National Library of Medicine (NLM) of the US.
PubMed provides free access to more than $30$M citations for biomedical literature, where each citation mainly contains the paper title, author information, abstract and semantic indices like MeSH (introduced in \S\ref{existing_kb}).
PubMed Central (PMC) includes full-texts of over $6$M biomedical articles in addition to the information provided in the PubMed citations.

More specific corpora are typically used for sub-domain BQA to filter out potential noise in larger corpora, e.g.: CORD-19 \citep{wang2020cord} for EPIC-QA Task A and Alzheimer's Disease Literature Corpus for QA4MRE-Alzheimer \citep{morante2012machine}.

\subsection{IR BQA Datasets}
\textit{\textbf{BioASQ Task B Phase A:}} BioASQ Task B is named ``Biomedical Semantic Question Answering" \citep{tsatsaronis2015overview} and contains two phases correspond to the IR and MRC BQA approaches in our BQA classification:
in the phase A (IR phase), systems retrieve relevant documents for the given question;
in the phase B (MRC phase), systems use gold standard relevant documents and ontological data to answer the given questions (discussed in \S\ref{mrcbqa}).
Specifically, for the given question, BioASQ phase A participants shall return the relevant:
1. Concepts from certain ontologies such as MeSH;
2. Articles from PubMed and text snippets within the articles;
3. RDF triples from the Linked Life Data\footnote{\url{http://linkedlifedata.com/}}.
Mean average precision (MAP) is used as a main metric for the BioASQ retrieval phase, with slight modifications for snippet retrieval.

\citet{lamurias2020generating} introduces \textit{\textbf{BiQA}}, an IR BQA dataset containing $7.4$k questions and $14.2$k relevant PubMed articles collected from online QA forums (Stack Exchange\footnote{\url{https://stackexchange.com/}} and Reddit\footnote{\url{https://www.reddit.com/}}).
They show that adding BiQA in the training set can marginally boost BioASQ Task B phase A test performance.

The Epidemic Question Answering (\textit{\textbf{EPIC-QA}}) challenges\footnote{\url{https://bionlp.nlm.nih.gov/epic_qa/}} are also formatted as IR BQA, where the participants return a ranked list of sentences from expert and consumer corpora to answer questions about the COVID-19 pandemic.

\textit{\textbf{HealthQA}} has $7.5$k manually annotated questions that can be answered by one of the $7.3$k web articles \citep{zhu2019hierarchical}.

The \textit{\textbf{TREC Genomics Tracks}} in 2006 and 2007 tackle the BQA with IR approach \citep{hersh2006trec, hersh2007trec, hersh2009trec}:
28 and 36 topic questions (e.g.: ``What [GENES] are genetically linked to alcoholism?") are released in 2006 and 2007, respectively, and the participating systems are required to retrieve passages from 162k full-text articles from Highwire Press\footnote{\url{https://www.highwirepress.com/}}.

IR BQA systems typically return a ranked list of documents as answers, and mean average precision (MAP) is usually used as the evaluation metric:
\[\text{AP}=\frac{1}{m}\sum_{k=1}^nP@k\times\text{rel}(k),\qquad \text{MAP}=\frac{1}{N}\sum_{q=1}^{N}\text{AP}_q\]
where $m$ denotes the number of gold-standard relevant documents and $n$ denotes the number of the returned documents. 
rel(k) is an indicator function that has the value 1 if the $k$-th document is relevant otherwise 0.

\subsection{IR BQA Methods}

\textit{\textbf{BioASQ Task B Phase A:}}
Wishart \citep{liu2013university} re-ranks and combines sentences from the retrieved documents to form the ideal answers for BioASQ task B phase B, and generate exact answers from the ideal answers according to the question type.
The USTB team \citep{jin2017multi} wins all batches in document, snippet and concept retrieval in BioASQ 5.
They use sequential dependence model \citep{bonnefoy2012social}, pseudo relevance feedback, fielded sequential dependence model \citep{zhiltsov2015fielded} and divergence from randomness model \citep{clinchant2009bridging}.
The AUEB team proposes a series of models \citep{brokos2018aueb, pappas2019aueb, pappas2020aueb} that win most of the Task B Phase A challenges since BioASQ 6.
At BioASQ 6, they \citep{brokos2018aueb} use the Position-Aware Convolutional Recurrent Relevance model \citep{hui2017pacrr} and the Deep Relevance Matching Model \citep{guo2016deep} for document retrieval, and use the Basic Bi-CNN model \citep{yin2016abcnn} for snippet retrieval. 
They win 3/5 and 5/5 batches for retrieving documents and snippets in BioASQ 6, respectively.
At BioASQ 7, they \citep{pappas2019aueb} combine the document and snippet retrieval system by modifying their BioASQ 6 system to also output the sentence-level (i.e.: snippet) relevance score in each document.
They win 4/5 and 4/5 batches for retrieving documents and snippets in BioASQ 7, respectively.
In BioASQ 8, They \citep{pappas2020aueb} continue to use this system and win 2/5 for document and 4/5 batches for snippet retrieval.

\textit{\textbf{HealthQA:}}
\citet{zhu2019hierarchical} propose Hierarchical Attention Retrieval, a ranking model for biomedical QA that uses a deep attention mechanism at word, sentence and document levels to compute a relevance score of the given query with each candidate document. 
With the proposed model, they achieve an MRR of 0.8788 on the HealthQA dataset.

\citet{zhou2007trec} win the \textit{\textbf{TREC 2006 Genomics Track}}, where they first identify query concepts and retrieve relevant documents with concept-level and word-level similarity measurements, and then extract the answers.
At \textit{\textbf{TREC 2007 Genomics Track}}, NLMinter \citep{demnerfushman2007combining} achieves the best performance.
NLMinter is an interactive retriever where the fusion retrieval results are boosted by the document relevance feedback, which is determined by expert PubMed search and occasional examination of the abstracts.

\subsection{Comments}
Though Classic BQA approaches usually contain a retrieval step, 
IR BQA is still considered as a distinct approach because the retrieved documents are directly used as answers,
and are evaluated by IR metrics.
Traditional retrieval methods like TF-IDF have been well-studied and ubiquitously used in IR BQA approach.
Future studies can focus more on PLM-based (re-)ranking methods \citep{chakraborty2020biomedbert, lin2020pretrained}, and how to better bridge the IR and MRC models.

\section{Machine Reading Comprehension BQA} \label{mrcbqa}
Machine Reading Comprehension (MRC) is a well-studied BQA task, where the models answer questions about given textual contexts. 
\textit{MRC BQA Datasets} are typically specialized in content and have predetermined answer format, so most \textit{MRC BQA Methods} developed on them are end-to-end neural models.

\subsection{MRC BQA Datasets}
Many MRC BQA datasets have been proposed, and we show an overview of them in Table \ref{tab:mrcbqa}.

\textit{\textbf{BioASQ Task B Phase B:}} it provide the largest and most widely used manually-annotated MRC BQA dataset:
Starting from 2013, BioASQ annotates about 500 test QA instances each year,
which will be included in the training set of the following years.
Currently, BioASQ 2020 consists of 3,243 training QA instances and at least 500 test instances.
Questions in BioASQ are typically scientific questions (\S\ref{scientific}).
There are 4 types of QA instances in BioASQ: factoid, list, yes/no and summary.
Factoid, list and yes/no instances have both \textit{exact} and \textit{ideal} answers:
\textit{Exact} answers are short answers that directly answer the questions, e.g.: single and multiple biomedical entities for factoid and list questions, respectively; ``yes" or ``no" for yes/no questions.
\textit{Ideal} answers are \textit{exact} answers written in complete sentences, e.g.: ``Yes, because [...]".
The main evaluation metrics for yes/no, factoid, list and summary questions are accuracy, MRR, mean F-score and manual score, respectively.
We show several examples of BioASQ instances in Table \ref{tab:bioasq_eg}.

\begin{table*}[ht!]
    \small
    \centering
    \setlength{\tabcolsep}{0.9mm}{
    \begin{tabular}{p{1.4cm}p{3.2cm}p{4cm}p{2cm}p{3.5cm}}
    \toprule
    \textbf{Type} & \textbf{Example Question} & \textbf{Example Context} & \textbf{Exact answer} & \textbf{Ideal answer} \\
    \midrule
    Yes / No & Is the protein Papilin secreted? & [...] and two genes encoding secreted extracellular matrix proteins, mig-6/papilin [...]. & Yes & Yes,  papilin is a secreted protein\\
    \midrule
    Factoid & Name synonym of Acrokeratosis paraneoplastica. & Acrokeratosis paraneoplastic (Bazex syndrome) is a rare, but [...] & Bazex syndrome & Acrokeratosis paraneoplastic (Bazex syndrome) is a rare [...]\\
    \midrule
    List & List Hemolytic Uremic Syndrome Triad. & Atypical hemolytic uremic syndrome (aHUS) is a rare disease characterized by the triad of [...]
    & anaemia, renal failure, thrombocytopenia
    & Hemolytic uremic syndrome (HUS) is a clinical syndrome characterized by [...] \\
    \midrule
    Summary & What is the effect of TRH on myocardial contractility? & Thyrotropin-releasing hormone (TRH) improved [...] & NA & TRH improves myocardial contractility\\
    \bottomrule
    \end{tabular}}
    \caption{Types of questions in BioASQ and respective examples}
    \label{tab:bioasq_eg}
\end{table*}

\begin{table*}[ht!]
    \small
    \centering
    \begin{tabular}{p{2.8cm}p{1.4cm}p{2cm}p{3.8cm}p{1cm}p{1.8cm}}
    \toprule
    \textbf{Dataset} & \textbf{Size} & \textbf{Metric} & \textbf{State-of-the-art (\%)} & \textbf{Content} & \textbf{Format} 
    \\ \midrule
    BioASQ Task B Phase B  \citep{tsatsaronis2015overview} & $3.7$k & F1, MRR, List F1, Manual & 90.3, 39.7, 52.3, 4.39/5  \citep{Nentidis2020overview} & Scientific & Y/N; Extraction; Generation   \\ \midrule
    Biomed-Cloze   \citep{dhingra2018simple} & $1$M & -- & -- & Scientific & Extraction   \\ \midrule
    BioMRC   \citep{pappas2020biomrc} & $812$k & Acc  & 80.06 (dev) / 79.97 (test) \footnotemark[1]   \citep{pappas2020biomrc} & Scientific & Extraction   \\ \midrule
    BioRead   \citep{pappas2018bioread} & $16.4$M & Acc & 47.06 (dev) / 51.52 (test)   \citep{pappas2018bioread} & Scientific & Extraction   \\ \midrule
    BMKC   \citep{kim2018pilot} & 473k (T); 370k (LS) & Acc & T: 85.5 (val) / 83.6 (test); LS: 80.1 (val) / 77.3 (test) \citep{kim2018pilot} & Scientific & Extraction   \\ \midrule
    CliCR   \citep{suster2018clicr} & $100$k & EM, F1 & 55.2, 59.8   \citep{pham2020machine} & Clinical & Extraction   \\ \midrule
    COVIDQA   \citep{tang2020rapidly} & $124$ & P@1, R@3, MRR & 30.6  \citep{chakravarti2020towards}, 47.7 \citep{su2020caire}, 41.5 \citep{tang2020rapidly} & Scientific & Extraction   \\ \midrule
    COVID-QA   \citep{moller2020covid} & $2$k & EM, F1 &  37.2, 64.7   \citep{reddy2020end}  & Scientific & Extraction   \\ \midrule
    EBMSummariser   \citep{molla2016corpus} & 456 & ROUGE-1,2,SU4 & 39.85, 24.50, 22.59 \citep{shafieibavani2016appraising} & Clinical & Generation   \\ \midrule
    emrQA   \citep{pampari2018emrqa} & $455$k & EM, F1 & 76.1, 81.7 \citep{rongali2020improved} & Clinical & Extraction   \\ \midrule
    MASH-QA   \citep{zhu2020question} & $34.8$k & EM, F1 & 29.49, 64.94   \citep{zhu2020question} & Consumer & Extraction   \\ \midrule
    MEDHOP   \citep{welbl2018constructing} & $2.5$k & Acc & 63.2   \citep{huo2020sentence} & Scientific & Multi-choice   \\ \midrule
    MEDIQA-AnS   \citep{savery2020question} & 552 (single); 156 (multi) & ROUGE-1,2,L; BLEU & Extractive\footnotemark[2]: 29, 15, 12, 9; Abstractive: 32, 12, 8, 9 \citep{savery2020question} & Consumer & Generation   \\ \midrule
    MEDIQA-QA    \citep{abacha2019overview} & $3$k & Acc, P, MRR & 79.49 \citep{he2020infusing}, 84.02 \citep{he2020infusing}, 96.22 \citep{pugaliya2019pentagon} & Consumer & Multi-choice   \\ \midrule
    ProcessBank   \citep{berant2014modeling} & 585 & Acc & 66.7 \citep{berant2014modeling} & Scientific & Multi-choice   \\ \midrule
    PubMedQA   \citep{jin2019pubmedqa} & $212$k & Acc, F1 & 68.08, 52.72 \citep{jin2019pubmedqa} & Scientific & Y/N   \\ \midrule
    QA4MRE-Alz   \citep{morante2012machine} & $40$ & $c@1$ & 76 \citep{bhaskar2012question} & Scientific & Multi-choice   \\ \midrule
    \end{tabular}
    \caption{An overview of the machine reading comprehension biomedical question answering datasets (listed in alphabetical order). 
    $^1$Results on BioMRC lite. 
    $^2$Results on single  document summarization. 
    }
    \label{tab:mrcbqa}
\end{table*}

Question Answering for Machine Reading Evaluation (\textit{\textbf{QA4MRE}}) holds a sub-task on machine reading of biomedical texts about Alzheimer's disease \citep{morante2012machine}.
This task provides only a test dataset with 40 QA instances, and each instance contains one question, one context and 5 answer choices.

\textit{\textbf{Cloze-style}} questions require the systems to predict the missing spans in contexts (e.g.: Q: ``Protein X suppresses immune systems by inducing \underline{\hbox to 8mm{}} of immune cells."; A: ``apoptosis").
There are many large-scale cloze-style MRC BQA datasets that are automatically constructed, such as CliCR, Biomed-Cloze, BioRead, BMKC and BioMRC.

\textit{\textbf{COVIDQA}} \citep{tang2020rapidly} is a QA dataset specifically designed for COVID-19.
It has 124 question-article pairs translated from the literature review page of Kaggle’s COVID-19 Open Research Dataset Challenge \footnote{\url{https://www.kaggle.com/allen-institute-for-ai/CORD-19-research-challenge/tasks}}, where relevant information for each category or subcategory in the review is presented.
\textit{\textbf{COIVD-QA}} \citep{moller2020covid} is another COVID-19 QA dataset with $2$k question-answer pair annotated by biomedical experts. 
The annotation is similar to that of SQuAD while the answers in COVID-QA tend to be longer as they generally come from longer texts. 

\citet{molla2011development, molla2016corpus} build the \textit{\textbf{EBMSummariser}}, a summarization dataset of 456 instances for EBM, from the Clinical Inquiries section of the Journal of Family Practice\footnote{\url{https://www.mdedge.com/familymedicine/clinical-inquiries}}: each instance contains a clinical question, a long ``bottom-line" answer, the answer's evidence quality and a short justification of the long answer.

\textit{\textbf{MASH-QA}} \citep{zhu2020question}, a dataset based on consumer health domain, is designed for extracting information from texts that span across a long document.
It utilizes long and comprehensive healthcare articles as context to answer generally non-factoid questions.
Different from the existing MRC datasets with short single-span answers, many answers in MASH-QA are several sentences long and excerpted from multiple parts or spans of the long context.

\textit{\textbf{MEDHOP}} \citep{welbl2018constructing} is a multi-hop MRC BQA dataset, where each instance contains a query of a subject and a predicate (e.g.: ``Leuprolide, interacts\_with, ?"), multiple relevant and linking documents and a set of answer options extracted from the documents.
Reasoning over multiple documents is required for the model to answer the question.

\textit{\textbf{MEDIQA-QA}} \citep{abacha2019overview} is the dataset of the QA subtask of MEDIQA 2019 shared task that has $400$ questions and $3$k associate answers. It is obtained by submitting medical questions to the consumer health QA system CHiQA then re-rank the answers by medical experts. The task of MEDIQA-QA dataset is to filter and improve the ranking of answers, making it a multi-choice QA task. \textit{\textbf{MEDIQA-AnS}} \citep{savery2020question}, on the other hand, is a summarization dataset. It provides extractive and abstractive versions of single and multi-document summary of the answer passages from MEDIQA-QA. 

\textit{\textbf{ProcessBank}} \citep{berant2014modeling} contains multi-choice questions along with relevant biological paragraphs. The paragraphs are annotated with "process", a directed graph $(\mathscr{T},\mathscr{A},\mathscr{E}_{tt},\mathscr{E}_{ta})$, where nodes $\mathscr{T}$ are token spans denoting the occurrence of events, nodes $\mathscr{A}$ are token spans denoting entities in the process, and the latter two are edges describing event relations and semantic roles respectively.

\citet{jin2019pubmedqa} build the \textit{\textbf{PubMedQA}} dataset from PubMed articles that use binary questions as titles (e.g.: ``Do preoperative statins reduce atrial fibrillation after coronary artery bypass grafting?") and have structured abstracts.
The conclusive parts of the abstracts are the long answers, while the main task of PubMedQA is to predict their short forms, i.e.: yes/no/maybe, using the abstracts without the conclusive parts as contexts.

\subsection{MRC BQA Methods}
In this section, we first introduce the top-performing systems in each year of the BioASQ challenge to reflect the landscape changes of MRC BQA methods.
We then briefly describe SoTA models of other surveyed MRC BQA datasets.

\textit{\textbf{BioASQ:}}
The first two BioASQ challenges \citep{partalas2013results, balikas2014results} use a Watson-motivated baseline \citep{weissenborn2013answering} that ensembles multiple scoring functions to rank the relevant concepts with type coercion to answer the given questions.

The Fudan system \citep{zhang2015fudan} of BioASQ 3B contains three major components:
1. A question analysis module that mainly extracts semantic answer types of questions;
2. Candidates generating by PubTator \citep{wei2013pubtator} and Stanford POS tools\footnote{\url{https://nlp.stanford.edu/software/tagger.shtml}};
3. Candidates ranking based on word frequency.
The SNU team \citep{choi2015snumedinfo} directly combines the retrieved relevant passages to generate the ideal answer and achieve state-of-the-art performance.
At BioASQ 4B: 
the HPI team \citep{schulze2016hpi} proposes an algorithm based on LexRank \citep{erkan2004lexrank} to generate ideal answers, which only uses biomedical named entities in the similarity function.
They win 1/5 batch in the ideal answer generation.
At BioASQ 5B:
the UNCC team \citep{bhandwaldar2018uncc} uses lexical chaining-based extractive summarization to achieve the highest ROUGE scores for ideal answer generation, with 0.7197 ROUGE-2 and 0.7141 ROUGE-SU4.

The CMU OAQA team describes a series of works for BioASQ \citep{yang2015learning, yang2016learning, chandu2017tackling}.
At BioASQ 3B, they propose a three-layered architecture \citep{yang2015learning} where:
the first layer contains domain-independent QA components such as input/output definition, intermediate data objects;
the second layer has implementations of biomedical domain-specific materials like UMLS and MetaMap \citep{aronson2001effective};
the third design layer is BioASQ-specific, including the end-to-end training and testing pipeline for the task.
The core components of the pipeline are the answer type prediction module and the candidate answer scoring module based on supervised learning.
At BioASQ 4B, they extend their BioASQ 3B system with general-purpose NLP annotators, machine-learning-based search result scoring, collective answer re-ranking and yes/no answer prediction \citep{yang2016learning}.
The CMU OAQA team is focused on ideal answer generation \citep{chandu2017tackling} at BioASQ 5B, using extractive summarization tools like Maximal Marginal Relevance \citep{carbonell1998use} and Sentence Compression \citep{filippova2015sentence} with biomedical ontologies such as UMLS and SNOMED-CT.
BioAMA \citep{sharma2018bioama} further improves the ROUGE score by 7\% for ideal answer generation than \citet{chandu2017tackling} by combining effective IR-based techniques and diversification of relevant snippets.

The Macquarie University team has participated in BioASQ 5-8 with the focus of ideal answer generation by extractive summarization for yes/no, factoid, list and summary questions \citep{molla2017macquarie, molla2018macquarie, molla2019classification, molla2020query}.
At BioASQ 5B, \citet{molla2017macquarie} observes that a trivial baseline that returns the top retrieved snippets as the ideal answer is hard to beat.
At BioASQ 6B, \citet{molla2018macquarie} show that using LSTM-based deep learning methods that predict the F1 ROUGE-SU4 score of an individual sentence and the ideal answer achieves the best results.
At BioASQ 7B, \citet{molla2019classification} observe that sentence-level classification task works better than regression task for finding the extractive summary sentences.

In recent years of BioASQ, transfer learning has gained increasing attention, where models are first pre-trained on large-scale general domain QA datasets or BQA datasets and then fine-tuned on the BioASQ training set.
\citet{wiese2017neural} achieve state-of-the-art performance on factoid questions and competitive performance on list questions by transferring the FastQA model pre-trained by SQuAD to BioASQ.
\citet{dhingra2018simple} show significant performance improvement over purely supervised learning by pre-training the GA-Reader \citep{dhingra2017gated} on an automatically generated large-scale cloze BQA dataset (\S\ref{automatic}) and then fine-tuning it on BioASQ.
\citet{du2018hierarchical, du2019hierarchical} have similar observations with transfer learning from the SQuAD dataset.
\citet{kang2020transferability} show transfer learning from NLI datasets also benefits BioASQ performance on yes/no (+5.59\%), factoid (+0.53\%) and list (+13.58\%) questions.
Generally, two main components are ubiquitously used in top-performing systems of the current BioASQ 8 challenge \citep{Nentidis2020overview}:
1. domain-specific pre-trained language models \citep{yoon2019pre}, such as BioBERT;
2. task-specific QA datasets that can (further) pre-train the used models, such as SQuAD for extractive QA and PubMedQA for yes/no QA.

In summary, with the introduction of large-scale MRC datasets like SQuAD \citep{rajpurkar2016squad, rajpurkar2018know}, a variety of neural MRC models have been proposed that incrementally improve the task performance, such as DCN \citep{xiong2016dynamic}, Bi-DAF \citep{seo2016bidirectional}, FastQA \citep{weissenborn2017making}.
Contextualized word embeddings pre-trained by language models (LM) like ELMo \citep{peters2018deep} and BERT \citep{devlin2019bert} show significant improvements on various NLP tasks including MRC.
Pre-trained LMs on biomedical corpora, such as BioELMo \citep{jin2019probing}, BioBERT \citep{lee2020biobert}, SciBERT \citep{beltagy2019scibert}, clinical BERT \citep{alsentzer2019publicly, huang2019clinicalbert} and PubMedBERT \citep{gu2020domain}, further improve their in-domain performance.
Probing experiments and analyses by \citet{jin2019probing} indicate that better encoding of biomedical entity-type and relational information leads to the superiority of domain-specific pre-trained embeddings.

Various methods have also been developed for other MRC BQA datasets.
Here we briefly discuss their representative SoTA methods as shown in Table \ref{tab:mrcbqa}.

\textit{\textbf{BioRead:}} \citet{pappas2018bioread} train the AOA Reader \citep{cui2017attention} on BioReadLite, which computes the mutual information between query and context, and places another attention layer over the document-level attention to achieve attended attention for the final prediction. They achieve the best accuracy of 0.5152. 
\textit{\textbf{BioMRC:}} It is the updated version of BioRead. \citet{pappas2020biomrc} use SciBERT \citep{beltagy2019scibert} and maximize scores of all mentions of each entity in the passage, achieving SoTA accuracy of 0.7997. 
\textit{\textbf{BMKC:}} Based on Attention Sum Reader architecture \citep{kadlec2016text}, \citet{kim2018pilot} present a new model that combines pre-trained knowledge and information of entity types. They also develop an ensemble method to integrate results from multiple independent models, which gets the accuracy of 0.836 on BMKC$\_$T and 0.773 on BMKC$\_$LS. 

\textit{\textbf{CliCR:}} \citet{pham2020machine} show that language models have better performance with systematic modification on cloze-type datasets. 
They replace \textit{@placeholder} with [MASK] and trains BioBERT \citep{lee2020biobert} on the modified dataset to obtain the SoTA EM of 0.552 and F1-score of 0.598. 

\textit{\textbf{COVIDQA:}} \citet{chakravarti2020towards} fine-tune pre-trained language models on the Natural Questions dataset \citep{kwiatkowski2019natural} with attention-over-attention strategy and attention density layer. They try its zero-shot transfer and achieve $P@1$ of 0.306. \citet{su2020caire} combine HLTC-MRQA \citep{su2019generalizing} with BioBERT to rank context sentences to get the evidence, and obtain $R@3$ of 0.477. \citet{tang2020rapidly} achieve MRR of 0.415 by fine-tuning T5 \citep{raffel2020exploring} on MS MARCO \citep{nogueira2020document}. 
\textit{\textbf{COVID-QA:}} \citet{reddy2020end} propose an example generation model for the training of MRC, and fine-tune RoBERTa-large \citep{liu2019roberta} on SQuAD2.0 \citep{rajpurkar2018know}, NQ and their generated training examples, which achieves EM of 0.372 and F1-score of 0.647.

\textit{\textbf{EBMSummariser:}} \citet{sarker2013approach} extract three sentences using hand-crafted features such as sentence length, position and question semantics for the EBMSummariser dataset, achieving ROUGE-L F-score of 0.168. \citet{shafieibavani2016appraising} utilize both UMLS and WordNet to summarise medical evidence for queries, and achieve ROUGE-1 of 0.3985, ROUGE-2 of 0.2450 and ROUGE-SU4 of 0.2259 on EBMSummariser. 

\textit{\textbf{emrQA:}} \citet{rongali2020improved} use rehearsal and elastic weight consolidation to improve domain-specific training, which can benefit the performance of models in both general domain and domain-specific tasks. They achieve EM of 0.761 and F1-score of 0.817. 

\textit{\textbf{MASH-QA:}} \citet{zhu2020question} propose MultiCo to select sentences across the long contexts to form answers. MultiCo combines a query-based sentence selection approach with a inter sentence attention mechanism, and achieves EM of 0.2949 and F1-score of 0.6494 on single-span MASH-QA dataset.

\textit{\textbf{MEDHOP:}} \citet{huo2020sentence} propose a Sentence-based Circular Reasoning approach which establishes a information path with sentence representation. They also implement a nested mechanism to systematically represent semantics, which improves the model performance significantly and achieves an accuracy of 0.632. 

\textit{\textbf{MEDIQA-AnS:}} \citet{savery2020question} train BART \citep{lewis2020bart} on the BioASQ data to achieve SOTA results. 
\textit{\textbf{MEDIQA-QA:}} \citet{he2020infusing} infuse disease knowledge into pre-trained language models like BERT and achieve accuracy of 0.7949 and precision of 0.8402. \citet{pugaliya2019pentagon} train their end-to-end system in a multi-task setting, and use the pretrained RQE and NLU modules to extract the best entailed questions and best candidate answers. They achieve MRR of 0.9622. 

\textit{\textbf{ProcessBank:}} \citet{berant2014modeling} first predict a structure representing the process in the given paragraph, then they map each question into queries and compare them with the predicted structure. They achieve the accuracy of 0.667. 

\textit{\textbf{PubMedQA:}} \citet{jin2019pubmedqa} take the multi-phase fine-tuning schedule with long answer as additional supervision, and achieve accuracy of 0.6808 and F1-score of 0.5272.

\subsection{Comments}
BioASQ is still the well-recognized benchmark and the ``go-to" dataset for MRC BQA because of its careful design, expert annotations, large size and highly active community.
Future models could explore developing pre-training methods that utilize richer biomedical knowledge than the raw texts (\S\ref{utilization}).
Additionally, collecting harder datasets / datasets that require other types of reasoning still remains an interesting future direction (\S\ref{difficulty}).

\section{Knowledge Base BQA} \label{kb}
KBQA (Knowledge Base QA) refers to answering questions using entities or relation information from knowledge bases \citep{fu2020survey}.
In biomedical domain, various large-scale biomedical KBs have been introduced, and one of their objectives is to assist with BQA.
Typically, one can convert natural language questions to SPARQL queries\footnote{\url{https://www.w3.org/TR/rdf-sparql-query/}} and use them to search the KBs for the answers.
In this section, we first introduce the \textit{existing knowledge bases} that have been used for KB BQA, and then introduce the \textit{KB BQA datasets} and \textit{KB BQA methods} developed on them.

\subsection{Existing Knowledge Bases} \label{existing_kb}
We define biomedical KBs as databases that describe biomedical entities and their relations, which can usually be stored by subject-predicate-object triples.
Biomedical KBs can be used for enhancing text representations \citep{jin2019probing,yuan2021coder,yuan-etal-2021-improving} and improving performances for BQA \citep{li2020towards} (not only KB BQA).
Substantial efforts have been made towards building biomedical KBs, including ontologies such as Medical Subject Headings (MeSH)\footnote{\url{https://www.ncbi.nlm.nih.gov/mesh/}} for biomedical text topics,
International Classification of Diseases (ICD)\footnote{\url{https://www.who.int/classifications/icd/}} for diseases and Systematized Nomenclature of Medicine Clinical Terms (SNOMED-CT, \citet{stearns2001snomed}) for medical terms.
The Unified Medical Language System (UMLS)\footnote{\url{https://www.nlm.nih.gov/research/umls}} is a metathesaurus that integrates nearly 200 different biomedical KBs like MeSH and ICD.
Biomedical KB is a big topic and we refer the interested readers to following references \citep{kamdar2020empirical, nicholson2020constructing}.



\subsection{KB BQA Datasets}
KB BQA datasets provide a list of biomedical questions and several biomedical KBs.
One should generate a SPARQL query for each question, and the answers are evaluated by query results.
We summary existing KB BQA datasets and show an overview of them in Table \ref{tab:kbbqa}.

\begin{table*}[ht!]
    \small
    \centering
    \begin{tabular}{p{3.5cm}p{1cm}p{1cm}p{2.5cm}p{1.5cm}p{1.5cm}}
    \toprule
    \textbf{Dataset} & \textbf{Size} & \textbf{Metric} & \textbf{State-of-the-art (\%)} & \textbf{Content} & \textbf{Format} \\
    \midrule
    Bioinformatics \citep{sima2021bio} & 30 & F1 & 60.0 \citep{sima2021bio} & Scientific & Generation \\ \midrule
    QALD-4 task 2 \citep{unger2014question} & 50 & F1 & 99.0 \citep{marginean2017question} & Consumer & Generation \\ 
    \bottomrule
    \end{tabular}
    \caption{An overview of KB BQA datasets (listed in alphabetical order).}
    \label{tab:kbbqa}
\end{table*}

\textit{\textbf{QALD-4 task 2:}} \citet{unger2014question} provides 50 natural language biomedical question and request SPARQL queries to retrieve answers from SIDER\footnote{\url{http://sideeffects.embl.de/}}, Drugbank\footnote{\url{https://www.drugbank.com/}} and Diseasome\footnote{\url{http://wifo5-03.informatik.uni-mannheim.de/diseasome/}}, where most questions require integrating knowledge from multiple databases to answer.
An example natural language question is ``Which genes are associated with breast cancer?'', and a possible query can be:
\begin{lstlisting}[captionpos=b, basicstyle=\ttfamily,frame=tb,xleftmargin=5em,xrightmargin=5em]
SELECT DISTINCT ?x
WHERE {
    diseases:1669 diseasome:associatedGene ?x . 
}
\end{lstlisting}

\textit{\textbf{Bioinformatics}} contains 30 biomedical queries with different complexity, and the database searching are restricted in Bgee\footnote{\url{https://bgee.org/sparql}} and OMA\footnote{\url{https://sparql.omabrowser.org/sparql}}.
The natural language questions include multiple concepts which leads to longer and more complicated SPARQL queries. 
    
\subsection{KB BQA Methods}
In this section, we introduce the well-performing KB BQA methods applies on \textit{\textbf{QALD-4 task 2}} and \textit{\textbf{Bioinformatics}}.


\textit{\textbf{QALD-4 task 2:}}
\citet{marginean2017question} wins QALD-4 task 2 by introducing the GFMed which is built with Grammatical Framework \citep{ranta2009gf} and Description Logic constructors, achieving 0.99 F1 on the test set. 
GFMed performs extraordinary well in QALD-4 task 2 since it is highly customized to this dataset.
CANaLI \citep{mazzeo2016question} designs an semantic automaton to parse the questions in specified form and achieves F1 of 0.92 on QALD-4 task 2. Questions not in specified form are ignored by CANaLI.
\citet{zhang2016joint} exploit KBs to find out candidate relation, type, entity + relation triple patterns in the questions.
They select and align triple patterns using integer linear programming and achieves F1 of 0.88 on QALD-4 task 2.
\citet{hamon2017querying} establish a complex pipeline to translate questions using existing NLP tools and semantic resources, and it achieves F1 of 0.85 on QALD-4 task 2.

\textit{\textbf{Bioinformatics:}}
\citet{sima2021bio} propose Bio-SODA which converts natural language questions into SPARQL queries without training data.
Bio-SODA generates a list of query graphs based on matched entities in the question, and rank the query graphs considering semantic and syntactic similarity, and node centrality.
Bio-SODA achieves F1 of 0.60 on QALD-4 task 2 and 0.60 on Bioinformatics.

Classic BQA systems also require natural language question translation systems to query KB.
\citet{rinaldi2004answering} adapt the general domain ExtrAns system \citep{molla2000extrans} to genomics domain.
They first convert the documents to Minimal Logical Forms (MLFs) and use them to construct a KB during the offline phase.
In the online QA phase, the system also converts the given question to MLFs by the same mechanism, and then get the answer by searching the built MLFs KB.
\citet{abacha2012medical, abacha2015means} propose MEANS for medical BQA which converts questions to SPARQL queries with a pipeline of classifying question types, finding Expected Answers Types, question simplification, medical entity recognition, extracting semantic relations and constructing SPARQL based on entities and semantic relations.
\citet{kim2013natural} introduce Linked Open Data Question Answering system to generate SPARQL queries for SNOMED-CT by predicate-argument relations from sentences.



\subsection{Comments}
Existing KB BQA datasets are limited by size, making it hard to train learning-based methods.
As a result, most top-performing KB BQA methods apply complex pipeline methods including entity extraction, relation extraction, entity alignment and entity typing to construct queries.
To leverage the potential of end-to-end deep learning methods, more labeled datasets are required for training a supervised Seq2seq question translation model.

\section{Question Entailment BQA} \label{qeapproach}
\citet{harabagiu2006methods} show that recognizing textual entailment can be used to enhance open QA systems.
The QE approach for BQA is essentially a nearest neighbor method that uses the answers of similar and already answered questions (e.g.: Frequently Asked Questions, FAQs) to answer the given question.
We will discuss three main components of this approach in this section: 
1. Models that can recognize similar questions, i.e.: \textit{QE BQA Methods};
2. datasets of similar \textit{Question-Question Pairs} for training QE models;
3. datasets of answered questions, i.e. \textit{Question-Answer Pairs}, that can be used to answer new questions.

\subsection{QE BQA Methods}
QE is formally defined by \citet{abacha2016recognizing} as: a question $\text{Q}_\text{A}$ entails a question $\text{Q}_\text{B}$ if every answer to $\text{Q}_\text{B}$ is also a correct answer to $\text{Q}_\text{A}$.
Natural language inference (NLI) is a relevant NLP task that predicts whether the relation of entailment, contradiction, or neutral holds between a pair of sentences.
In the general domain, predicting question-question similarity is an active research area with potential applications in question recommendation and community question answering \citep{nakov2016semeval, nakov2017semeval}.

\citet{luo2015simq} propose the SimQ system to retrieve similar consumer health questions on the Web using UMLS-annotated semantic and AQUA-parsed \citep{campbell2002transformational} syntactic features of the questions. 
CMU OAQA \citep{wang2017cmu} use a dual entailment approach with bidirectional recurrent neural networks (RNN) and attention mechanism to predict question similarity;
\citet{abacha2019question} use feature-based logistic regression classifier and deep learning models that pass the concatenation of two question embeddings to multiple ReLU layers \citep{nair2010rectified} for recognizing QE;
\citet{Zhu2019PANLPAM} fine-tune pre-trained language models to classify question pairs and conduct transfer learning from NLI to boost the performance.

\subsection{Question-Question Pairs}
Training QE models needs datasets of question pairs annotated with entailment (similarity) labels.
Towards this end, \citet{abacha2016recognizing} introduce the clinical-QE dataset\footnote{\url{https://github.com/abachaa/RQE_Data_AMIA2016}} that contains over $8$k training biomedical question pairs with similarity labels.
The questions are from clinical questions collected by \citet{ely2000taxonomy}, Consumer Health Questions and FAQs from NLM and NIH websites, respectively.
This dataset is also used as the RQE dataset in the MEDIQA challenge with slight changes.
\citet{sun2020analysis, poliak2020collecting, zhang2020cough} build question-question relevance datasets along with their FAQ datasets on COVID-19.

In general, only limited efforts have been made to build biomedical QE datasets, which results in the lack of training instances.
Many works instead consider a transfer learning approach to pre-train the QE models on other text pair tasks, including biomedical natural language inference (NLI) datasets like MedNLI \citep{romanov2018lessons}, general domain QE datasets like SemEval-cQA \citep{nakov2016semeval, nakov2017semeval}, and general domain NLI datasets like SNLI \citep{bowman2015large} and MultiNLI \citep{williams2018broad}.

\subsection{Question-Answer Pairs}
QE approach relies heavily on large databases of question-answer pairs with high quality to answer unseen questions.
For this, \citet{abacha2019question} build MedQuAD, a collection of 47,457 question-answer pairs from trusted websites e.g.: \url{https://www.cancer.gov/}.
Using MedQuAD for BQA can protect users from misleading and harmful health information since most answers are well-curated.
Moreover, several FAQ datasets have been introduced for answering COVID-19 related questions \citep{sun2020analysis, poliak2020collecting, zhang2020cough}.
However, since expert-curated answers are expensive to collect, such question-answer pair datasets might be limited in size.
Online health and QA communities like WebMD\footnote{\url{https://www.webmd.com/}}, Quora\footnote{\url{https://www.quora.com/}} provide large amounts of QA pairs, and many large-scale BQA datasets have been built using online doctor-patient QA data \citep{zhang2017chinese, zhang2018multi, he2019applying, tian2019chimed}.
These materials can be potentially used in QE approach, but the quality of user-provided answers should be carefully checked.

\subsection{Comments}
The most important components for the QE BQA approach are the datasets of question-question (Q-Q) and question-answer (Q-A) pairs.
However, these datasets are currently limited in scale or quality.
To tackle this issue, methods for automatically collecting large-scale Q-Q and Q-A datasets with high quality should be explored (\S\ref{automatic}).

\section{Challenges and Future Directions} \label{challenge}
In this section, we discuss the challenges identified in \S \ref{intro} with the surveyed works.
We also propose some interesting and potential future directions to explore.
\S\ref{automatic} and \S\ref{difficulty} involve dataset scale and difficulty, respectively.
In \S\ref{vqa}, we discuss visual BQA, which is an active and novel research field that is gaining more and more attention.
We explain why domain knowledge is not fully utilized in BQA and how the fusion of different BQA approaches can potentially solve it in \S\ref{utilization}.
In \S\ref{explainability}, we study different forms of explainability of BQA systems.
In \S\ref{eval}, we discuss two main issues of BQA system evaluation: qualitatively, what parts of the systems are evaluated and quantitatively, how they are evaluated.
Last but not the least, we discuss the fairness and bias issues of BQA in \S\ref{fairness}.

\subsection{Dataset Collection} \label{automatic}
Annotating large-scale BQA datasets is prohibitively expensive since it requires intensive expert involvement.
As a result, the majority of BQA datasets are automatically collected.
There are 3 main approaches for automatically collecting BQA datasets: 
question generation, cloze generation and exploiting existing QA pairs.

\paragraph{\textbf{Question Generation:}}
Question generation (QG) can automatically generate questions from given contexts \citep{du2017learning}, which can be utilized to build QA datasets.
\citet{yue2020cliniqg4qa} apply the QG approach to synthesize clinical QA datasets without human annotations, and show that the generated datasets can be used to improve BQA models on new contexts.

\paragraph{\textbf{Cloze Generation:}} 
Cloze-type QA is the task of predicting a removed word or phrase in a sentence, usually using a detailed context.
Several biomedical QA datasets have been collected by cloze generation, such as CliCR, BioRead, BMKC and BioMRC.
Most of them follow a similar process to \citet{hermann2015teaching} for generating the CNN and Daily Mail datasets: 
1. Recognizing biomedical named entities appearing in both summary sentences (e.g. article titles, article learning points) and their detailed contexts (e.g. article abstracts) with tools like MetaMap; 
2. Masking the recognized named entities in the summary sentences;
3. The task is to predict the named entities using the masked summary sentences and the contexts.
The generated datasets are typically large-scale (ranging from $100$k to $16.4$M instances) and thus can be used for pre-training \citep{dhingra2018simple} or as a task itself \citep{pappas2018bioread, kim2018pilot, pappas2020biomrc}.
When used for pre-training, cloze-type QA is actually a special type of language modeling that predicts only biomedical entities and is conditioned on the corresponding contexts.

\paragraph{\textbf{Exploiting Existing QA Pairs:}}
Another widely used approach for dataset collection is to exploit naturally existing QA Pairs, or exploiting domain-specific corpora structures.
Biomedical question-answer pairs can be found in a variety of contexts:
For example,
PubMedQA \citep{jin2019pubmedqa} collects citations in PubMed whose titles are questions, and uses the conclusive part of the abstracts as ideal answers.
MedQA \citep{zhang2018medical}, HEAD-QA \citep{vilares2019head} and NLPEC \citep{li2020towards} are built from QA pairs in medical examinations.
MedQuAD collects expert-curated FAQs from trusted medical websites.
cMedQA \citep{zhang2017chinese, zhang2018multi} and ChiMed \citep{tian2019chimed} exploit doctor-patient QA pairs on online healthcare communities.


\subsection{Dataset Difficulty} \label{difficulty}
BQA datasets should be difficult to evaluate the non-trivial reasoning abilities of BQA systems.
In this section, we discuss three types of advanced reasoning abilities: answerability, multi-hop and numeric reasoning.

\paragraph{\textbf{Answerability reasoning:}}
Almost all current BQA datasets and methods focus on answerable questions.
However, not all questions in biomedicine are answerable, the fact that only answerable questions are evaluated can be exploited by BQA systems to get high performance without the expected reasoning process (e.g. by identifying the only text snippet in the context that is consistent with the expected lexical answer type).
In general domain, a strong neural baseline drops from $86$\% F1 on SQuAD v1.1 to $66$\% F1 after unanswerable questions are added \citep{rajpurkar2018know}.
It remains an interesting direction to add unanswerable questions in BQA datasets and test the robustness of BQA systems under such settings.

\paragraph{\textbf{Multi-hop reasoning:}}
Answering real biomedical questions often requires multi-hop reasoning.
For example, doctors might ask ``What tests shall we conduct for patients with [certain symptoms]?".
To answer this question, models must: 1. infer the possible diseases which the patient might have, and 2. find the available tests that are needed for the differential diagnosis.
However, to the best of our knowledge, only MEDHOP evaluates multi-hop reasoning abilities, while almost all other BQA datasets focus on singe-hop reasoning.

\paragraph{\textbf{Numeric reasoning:}}
Numbers are widely used in modern biomedical literature, so understanding the numeric contents in texts is necessary to correctly answer non-trivial scientific questions.
\citet{jin2019pubmedqa} show that nearly all questions from PubMed article titles require quantitative reasoning to answer.
While about 3/4 of them have text descriptions of the statistics, e.g. ``significant differences", about 1/5 only have numbers.
For this, \citet{wu2021bionumqa} re-annotate the PubMedQA and BioASQ datasets with numerical facts as answers, and show that adding numerical encoding scheme improves BioBERT performance on their dataset.
However, most current BQA systems treat numbers as text tokens and do not have specific modules to process them.

\subsection{Biomedical VQA} \label{vqa}
Biomedical VQA is a novel BQA task.
In biomedical VQA, questions are asked about images, which are ubiquitously used and play a vital role in clinical decision making. 
Since manual interpretation of medical images is time-consuming and error-prone, automatically answering natural language questions about medical images can be very helpful.
VQA is a novel variant of QA task that require both NLP techniques for question understanding and Computer Vision (CV) techniques for image representation.
General VQA is an active research area and there has been many recent survey articles \citep{gupta2017survey, wu2017visual, srivastava2019visual}.
Here, we mainly focus on \textit{Biomedical VQA Datasets} and their corresponding \textit{Multi-Modal Methods} that fuse NLP and CV methods.

\begin{table*}[ht!]
    \small
    \centering
    \begin{tabular}{p{2cm}p{1cm}p{3.5cm}p{3.5cm}p{1cm}p{1.2cm}}
    \toprule
    \textbf{Dataset} & \textbf{Size} & \textbf{Metric} & \textbf{State-of-the-art (\%)} & \textbf{Content} & \textbf{Format} \\
    \midrule
    PathVQA  \citep{he2020pathvqa} & $32.8$k & Acc, BLEU-1, BLEU-2, BLEU-3 & 68.2, 32.4, 22.8, 17.4 \citep{he2020pathvqa}  & Clinical & Generation  \\ \midrule
    VQA-Med  \citep{abacha2019vqa} & $15.3$k & Acc, BLEU  & 64.0, 65.9  \citep{ren2020cgmvqa} & Clinical & Generation  \\ \midrule
    VQA-Rad  \citep{lau2018dataset} & $3.5$k & Acc & 60.0 (open) / 79.3 (close) \citep{zhan2020medical}  & Clinical & Generation \\ 
    \bottomrule
    \end{tabular}
    \caption{An overview of biomedical VQA datasets (listed in alphabetical order).}
    \label{tab:vqa}
\end{table*}

\begin{figure}[ht!]
    \centering
    {\includegraphics[width=0.72\linewidth]{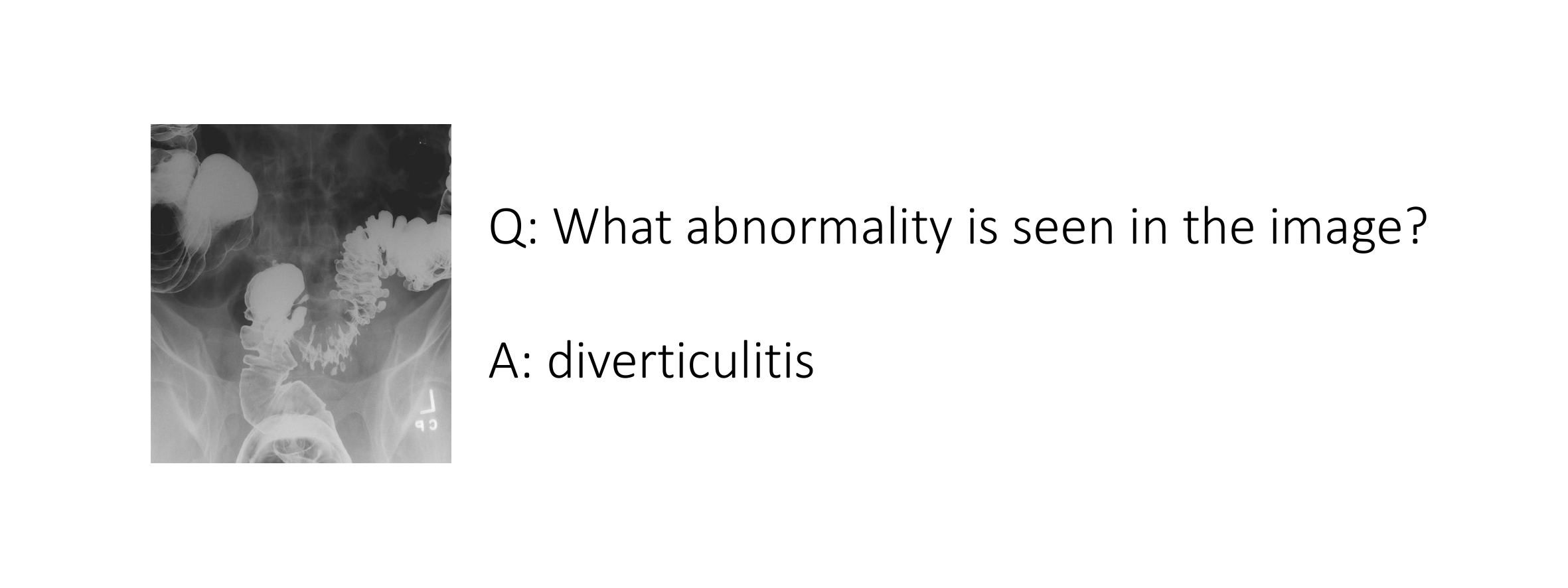}}
    \caption{An instance of VQA-Med.}
    \label{fig:vqa}
\end{figure}

\paragraph{{Biomedical VQA Datasets}}
We show an overview of biomedical VQA datasets in Table \ref{tab:vqa} and an instance in Figure \ref{fig:vqa}.
\textbf{VQA-Rad} \citep{lau2018dataset} is the first VQA dataset for radiology. 
It contains $3.5$K QA pairs that are annotated by clinicians and the images are from MedPix\footnote{\url{https://medpix.nlm.nih.gov/home}}. 
Questions in VQA-Rad are classified into modality, plane, organ system, abnormality, object/condition presence, etc.
Answer formats include multi-choices and generations in VQA-Rad.
The \textbf{VQA-Med} \citep{abacha2019vqa} dataset is automatically constructed using captions of the radiology images. 
There are $15.3$K questions in VQA-Med that are restricted to be about only one element and should be answerable from the corresponding images. 
VQA-Med concentrates on the most common questions in radiology, which include categories of modality, plane, organ system and abnormality. 
Yes/no and WH-questions are included in VQA-Med. 
\textbf{PathVQA} \citep{he2020pathvqa} semi-automatically extracts $32.8$K pathology images and generates the corresponding answers from textbooks.

\paragraph{{Multi-Modal Methods}}
Typically, for biomedical VQA, images and texts are separately encoded, and a multi-modal pooling mechanism is often used to obtain the mixed representations for generating the answers.
\textit{\textbf{Image encoders:}} VGGNet \citep{simonyan2014very} and ResNet \citep{he2016deep} are commonly used for image feature extraction.
\citet{yan2019zhejiang, ren2020cgmvqa} adopt global average pooling \citep{lin2013network} on VGGNet for image encoding to prevent over-fitting on small datasets.
To overcome data limitation of images, \citet{nguyen2019overcoming} apply model-agnostic meta-learning \citep{finn2017model} and convolutional denoising auto-encoder \citep{masci2011stacked} to initialize CNN layers on VQA-Rad, and achieve 43.9 and 75.1 accuracy on open-ended and close-ended questions, respectively.
\textit{\textbf{Text encoders:}} Questions are usually encoded by a recurrent network or a pre-trained language model similar other BQA approaches. 
The co-attention mechanism is used for finding important words and regions to enhance both textual and visual representation.
Stacked attention networks \citep{yang2016stacked} use text representation to query visual representation multiple times for obtaining multi-step reasoning.
\textit{\textbf{Multi-modal pooling:}} it is crucial to combine features from the visual and textual encoders. Direct concatenation of them can serve as a baseline. 
Multi-modal compact bilinear pooling \citep{fukui2016multimodal}, multi-modal factorized bilinear pooling \citep{yu2017multi} and multi-modal factorized high-order pooling \citep{yu2018beyond} are often used for feature fusion in VQA.
Recently, several multi-modal pre-trained models have been proposed that use transformers \citep{li2019visualbert, tan-bansal-2019-lxmert} to generate visual and textual representations in the general domain.
Similarly, \citet{ren2020cgmvqa} introduce the CGMVQA model that feeds VGGNet and word embedding features into a single transformer for classification or generation on VQA-Med, achieving the accuracy of 0.640 and BLEU of 0.659.

\subsection{Domain knowledge Utilization} \label{utilization}
There are a variety of biomedical domain-specific materials and tools that can be used in BQA, including:
1. Large-scale corpora like PubMed and PMC that contain millions of freely available biomedical articles;
2. Various biomedical KBs like UMLS and DrugBank;
3. Many domain-specific NLP tools, e.g.: MetaMap and SemRep for identifying biomedical entities and relations, respectively.
Each kind of resource has its advantages and disadvantages:
Biomedical raw textual resources are extremely large-scale, but their quality cannot be assured.
Specific textual resources, e.g.: FAQs from NLM websites, are regularly maintained and thus of high quality, but they are limited in scale since maintaining and collecting them is expensive.
KBs have high quality and intensive knowledge, but most of them are sparse and incomplete.

However, the abovementioned resources have not been fully utilized by current BQA systems.
As shown in Table  \ref{tab:approaches}, different BQA approaches use only one or two different types of resources, but not all of them.
For example, IR, MRC and QE BQA systems typically use textual data, while KB BQA systems mainly use the KBs.
Biomedical NLP tools are mostly used in classic BQA systems.
Since each resource only encodes certain types of biomedical knowledge, only by fusing different BQA approaches can systems fully utilize the domain knowledge.

\begin{table}[ht!]
    \small
    \centering
    \begin{tabular}{p{4.5cm}p{3cm}p{1cm}p{1.5cm}p{3cm}}
    \toprule
    \textbf{BQA Approach} & \textbf{Texts} & \textbf{Images} & \textbf{KBs} & \textbf{BioNLP tools} \\
    \midrule
    Information Retrieval & Document collections & -- & -- & -- \\
    Machine Reading Comprehension & Raw documents (Contexts) & -- & -- & -- \\
    Knowledge Base & -- & -- & \checkmark & Used for KB construction \\
    Question Entailment & Existing FAQs & -- & -- & -- \\
    Visual Question Answering & -- & \checkmark & -- & -- \\
    Classic & Document collections & -- & Ontologies & \checkmark \\
    \bottomrule
    \end{tabular}
    \caption{Utilized domain knowledge by different BQA approaches.}
    \label{tab:approaches}
\end{table}

The KMQA model \citep{li2020towards} combines the IR-MRC approach and the KB approach by using co-attention mechanism and a novel knowledge acquisition algorithm.
Their ablation experiments show that only using texts achieves 64.6\% accuracy on the development set; only using knowledge bases achieves 45.3\%; using both texts and knowledge bases achieves 71.1\%. 
This shows the effectiveness of the fusion of different BQA approaches.
However, it still remains an underexplored area.

\subsection{Answer Explainability} \label{explainability}
Explainability is a vital property of healthcare applications.
An ideal BQA model should not only have high accuracy in predicting the exact answers, but also be able to provide explanations or evidence for giving such answers.
This improves the answer reliability and enables further fact checking.

Each BQA approach has its intrinsic way for answer explanation:
for the IR approach, the retrieved documents can be considered as evidence;
for the KB approach, the reasoning paths in the KBs provide explainability;
for the QE approach, users are directly pointed to similar questions that have already been answered.
Though controversial \citep{jain2019attention, wiegreffe2019attention}, the attention mechanism \citep{bahdanau2014neural} that is ubiquitously used in modern BQA systems provide at least some level of explainability.
To evaluate explicit answer explanations, Phase B of BioASQ challenges also require the participants to submit ``ideal answers", i.e. answers that include both the exact answers and explanations, in addition to exact answers.

\citet{zhang2019multi} generate explanations with medical KB paths that link the entities extracted from consumer health questions and doctors' answers.
The path representations are learned by a translation-based method and the weights of reasoning paths for specific QA pairs are generated by a hierarchical attention network.
They also use the entity figures return from Google for better entity representation and consumer understanding.
\citet{liu2020interpretable} presents the MURKE model to solve HEAD-QA which iteratively select the most relative document to reformulate the question, where the series of modified questions can be considered as an interpretable reasoning chain.

\subsection{Evaluation Issues} \label{eval}
\paragraph{\textbf{Modular evaluation:}}
Most current evaluations are modular because they only evaluate certain parts of the full BQA system, e.g. for the IR-MRC BQA approach, BioASQ Task B phase A only evaluates the IR methods and the Phase B provides gold standard contexts and only evaluates the MRC methods.
The majority of BioASQ teams only participate in one phase \citep{Nentidis2020overview}.
However, in real settings: 1. it's impossible to have the relevant documents; 2. state-of-the-art MRC BQA systems might not perform well given non-perfect retrieved documents \citep{lin2020pretrained}.
As a result, closing the gap between system modules by combining the evaluations is vital to test the real utility of BQA systems.

In the general domain, \citet{chen2017reading} propose the Machine Reading at Scale task for the complete IR-MRC QA evaluation.
They show that the performance of a complete QA system that reads all Wikipedia might have a large drop compared to its MRC component that reads only the gold standard contexts, e.g.: from 69.5\% EM to 27.1\% on the development set of SQuAD.
In the biomedical domain, many datasets that only contain questions and answers have been proposed.
We list these datasets in Table \ref{tab:openbqa}, most of which are related to Consumer health (5/10) or Examination (4/10), because their dataset sources typically have no supporting materials for the answers.
It should be noted that other types of BQA datasets can also be converted to such datasets by removing the supporting materials (document collections, contexts, FAQs etc).

\begin{table*}[ht!]
    \small
    \centering
    \begin{tabular}{p{2cm}p{1cm}p{1.5cm}p{5.2cm}p{1.5cm}p{1.6cm}}
    \toprule
    \textbf{Dataset} & \textbf{Size} & \textbf{Metric} & \textbf{State-of-the-art (\%)} & \textbf{Content} & \textbf{Format}
    \\ \midrule
    ChiMed  \citep{tian2019chimed} & $24.9$k & Acc &  98.32 (rel.) / 84.24 (adopt.) \citep{tian2019chimed} & Consumer & Multi-choice  \\ \midrule
    cMedQA  \citep{zhang2017chinese} & $54$k & P@1 & 65.35 (dev) / 64.75 (test) \citep{zhang2017chinese} & Consumer & Multi-choice  \\ \midrule
    cMedQA v2  \citep{zhang2018multi} & $108$k & P@1 & 72.1 (dev) / 72.1 (test) \citep{zhang2018multi} & Consumer & Multi-choice  \\ \midrule
    HEAD-QA  \citep{vilares2019head} & $6.8$k & Acc & 44.4 (supervised) / 46.7 (unsupervised) \citep{liu2020interpretable} & Examination & Multi-choice  \\ \midrule
    LiveQA-Med  \citep{abacha2017overview} & $738$ & avgScore & 82.7 \citep{abacha2016recognizing} & Consumer & Generation  \\ \midrule
    MedQA  \citep{zhang2018medical} & $235$k & Acc & 75.8 (dev) / 75.3 (test) \citep{zhang2018medical} & Examination & Multi-choice  \\ \midrule
    MEDQA  \citep{jin2020disease} & $61$k & Acc &  MC: 69.3 (dev) / 70.1 (test); TW: 42.2 (dev) / 42.0 (test); US: 36.1 (dev) / 36.7 (test) \citep{jin2020disease} & Examination & Multi-choice  \\ \midrule
    NLPEC   \citep{li2020towards} & $2.1$k & Acc & 71.1 (dev) / 61.8 (test) \citep{li2020towards} & Examination & Multi-choice  \\ \midrule
    webMedQA  \citep{he2019applying} & $63$k & P@1, MAP & 66.0, 79.5  \citep{he2019applying} & Consumer & Multi-choice \\
    \bottomrule
    \end{tabular}
    \caption{An overview of the BQA datasets that contain no supporting materials (listed in alphabetical order).}
    \label{tab:openbqa}
\end{table*}

Olelo \citep{kraus2017olelo} and Bio-AnswerFinder \citep{ozyurt2020bio} are complete QA systems that participate in the BioASQ challenge.
Olelo is proposed as an integrated web application for QA-based exploration of biomedical literature.
For each user question, Olelo uses the HPI system at BioASQ 2016 (\citealt{schulze2016hpi}, described in \S\ref{mrcbqa}) to retrieve relevant abstracts and return the answers, as well as the entity-supported summarizations of the abstracts \citep{schulze2016entity}.
Bio-AnswerFinder uses iterative document retrieval by LSTM-enhanced keyword queries and BERT-based answer ranking.
The system performance is comparable to a BioASQ 5 SoTA MRC system for factoid questions (38.1\% v.s. 40.5\% MRR, \citet{wiese2017neural}), but is still lower than BioBERT (38.1\%, 48.3\%).
The baselines of ChiMed, cMedQA and webMedQA use answer matching models without explicit supporting materials, and the baselines provided by HEAD-QA, MedQA and MEDQA are basically combined IR-MRC approach (\S\ref{irbqa} and \S\ref{mrcbqa}). 
Since most of the current BQA evaluations only focus on the MRC, tasks that involve both retrieving relevant contents and comprehending over them should be further explored.

\paragraph{\textbf{Evaluation metrics:}} 
In extractive and generative BQA, current metrics do not consider the synonyms of biomedical concepts.
For example, if the ground truth answer is ``kidney diseases", ``renal diseases" should conceptually be an exact match and ``renal insufficiency" should be rated as relevant. 
However, if we use EM in practice, both ``renal diseases" and ``renal insufficiency" have a score of 0;
if we use F1, BLEU or ROGUE, ``renal diseases" is only a partial match and `renal insufficiency" has a score of 0.
\citet{wiese2017neural} report that their model predictions of 10 among 33 analyzed questions are synonyms to the gold standard answers,
but are not counted as right in BioASQ evaluation.

There are two potential approaches to solve this problem:
1. From the annotation side, we can manually annotate more gold standard answers. This approach is expensive but the quality is guaranteed;
2. From the metrics side, it's worth exploring to infuse domain knowledge (e.g.: UMLS ontologies) into current evaluation metrics.
For example, to consider the rich concept synonyms in biomedicine during evaluation, \citet{suster2018clicr} also evaluates QA models by a cosine similarity metric between the mean word vectors of the ground truth and the predicted answer.

\subsection{Fairness and Bias} \label{fairness}
Fairness and bias are serious and vital issues in machine learning.
One cannot be more cautious in removing all potential biases, e.g.: racial and gender biases, when developing healthcare applications like BQA.
Here we discuss the fairness and bias issues of BQA from the NLP and the biomedical side.
From the NLP side: Word embeddings \citep{mikolov2013distributed} are ubiquitously used in NLP models, but such embeddings result in biased analogies like: ``man" is to ``doctor" as ``woman" is to ``nurse".
Similar trends have been observed \citep{kurita2019measuring} in contextualized word representations like BERT.
From the biomedical side, since most current BQA models learn from historically collected data (e.g.: EMRs, scientific literature), populations that have experienced structural biases in the past might be vulnerable under incorrect predictions \citep{rajkomar2018ensuring}.

Some works have been done in general NLP and biomedical machine learning domains, but only a little progress has been made in the BQA domain, and the majority of them study non-English BQA:
Unlike English BQA, non-English BQA suffers additional challenges mainly from the lack of domain-specific resources: much less scientific literature are available in non-English languages; general NLP tools are scarce for non-English languages, let alone biomedical domain-specific ones.
\citet{jacquemart2003towards, delbecque2005indexing} present preliminary studies of BQA in French.
\citet{olvera2011multilingual} evaluate multilingual QA system HONQA and find that English questions are answered much better than French and Italian. 
Researchers also introduce multi-lingual BQA datasets for low-resource languages: \citet{zhang2017chinese, zhang2018medical, zhang2018multi, tian2019chimed, he2019applying} for Chinese, \citet{vilares2019head} for Spanish and \citet{veisi2020persian} for Persian. 

However, current works are far from enough and our community should seriously take fairness and bias issues into account when introducing new BQA datasets and algorithms in the future.



\bibliographystyle{ACM-Reference-Format}
\bibliography{manuscript}

\end{document}